**Crowd Safety Manager:**
**Towards Data-Driven Active Decision Support for Planning and Control of Crowd Events**

Panchamy Krishnakumari
Department of Transport & Planning
Delft University of Technology, Delft, The Netherlands
Email: p.k.krishnakumari@tudelft.nl

Sascha Hoogendoorn-Lanser
Mobility Innovation Centre Delft, Impact & Innovation Center
Delft University of Technology, Delft, The Netherlands
Email: s.hoogendoorn-lanser@tudelft.nl

Jeroen Steenbakkers
Argaleo B.V.
Email: jeroen@argaleo.com

Serge Hoogendoorn (corresponding author)
Department of Transport & Planning
Delft University of Technology, Delft, The Netherlands
Email: s.p.hoogendoorn@tudelft.nl

Word Count: 6900 words + 2 tables (250 words per table) = 7,400 words

Submitted 31/7/2023

## ABSTRACT

This paper presents novel technology and methodology aimed at enhancing crowd management in both the planning and operational phases. The approach encompasses innovative data collection techniques, data integration, and visualization using a 3D Digital Twin, along with the incorporation of artificial intelligence (AI) tools for risk identification. The paper introduces the Bowtie model, a comprehensive framework designed to assess and predict risk levels. The model combines objective estimations and predictions, such as traffic flow operations and crowdedness levels, with various aggravating factors like weather conditions, sentiments, and the purpose of visitors, to evaluate the expected risk of incidents.

The proposed framework is applied to the Crowd Safety Manager project in Scheveningen, where the DigiTwin is developed based on a wealth of real-time data sources. One noteworthy data source is Resono, offering insights into the number of visitors and their movements, leveraging a mobile phone panel of over 2 million users in the Netherlands.

Particular attention is given to the left-hand side of the Bowtie, which includes state estimation, prediction, and forecasting. Notably, the focus is on generating multi-day ahead forecasts for event-planning purposes using Resono data. Advanced machine learning techniques, including the XGBoost framework, are compared, with XGBoost demonstrating the most accurate forecasts. The results indicate that the predictions are adequately accurate. However, certain locations may benefit from additional input data to further enhance prediction quality. Despite these limitations, this work contributes to a more effective crowd management system and opens avenues for further advancements in this critical field.

## BACKGROUND

Situations where many people gather are increasingly frequent: crowds may occur on a regular basis at busy transfer points, in shopping streets, tourists visiting a city centre during the holiday season, at the beachside when weather conditions are good, or on an incidental basis, for instance in case of a demonstration, a planned event in the city, or a music festival. The risks associated with such conditions vary from nuisance and rioting [1] to deadly crushing caused by overcrowding at pinch-points or due to turbulence [2].

While substantial research has been done on modelling crowds and predicting pedestrian traffic flow operations, science has not been able yet to provide methods allowing to accurately assess, predict and forecast risks associated with crowding. This is first due to the complex nature of the process of crowding and the non-trivial relation between objective crowding (i.e., the number of people in an area) and its context (e.g., purpose of people in the crowd, sentiments, crowd control interventions, and external conditions such as weather). Secondly, the dynamics of the system are strongly non-linear and have chaotic features, making the occurrence of high-risk situations highly ill-predicable [2]. This complexity is very hard to capture in traditional physics-inspired crowd flow models: a data-driven approach is needed to unravel such complex system dynamics and emergent phenomena. Thirdly, the lack of adequate sources of data has further complicated analysis, risk assessment and prediction, and consequently the ability to deploy effective interventions and policies. As a final point, decision theories routinely assume that a decision maker has time and attention to screen and evaluate alternatives and select the preferred option [3], or that the decisions are part of professional training or daily routines [4]. However, emergent situations create unprecedented and unplanned for decision situations where time is scarce and information is volatile, often pushing experts beyond their experience [5]. The key to effective response is access to relevant, timely and actionable information [6].

The lack of adequate methods and decision support tools implies a major issue for the various stakeholders involved (event organisers, police, emergency services, city authorities, etc.) since decision makers must rely on their past experiences, and fragmented subjective observations 'on the ground'. Although in the past several dashboard tools have been implemented to provide more objective data to decision makers, issues like information quality, usefulness, timeliness, trust, and the relation of the data to the actual risks arise [7]. Moreover, the form of the information needs to align with the decision maker's cognitive state.

Summarizing, there is a strong need for (more) accurate and timely data collection, state (crowding) estimation and prediction, and forecasting tools, as well as for advanced support systems that can assess the risk of the current or future situation in relation to all related factors on which risk of a calamity depends *and* provide decision makers with information befitting their information needs. In developing such a system, we believe that many crises can be prevented, mitigated, or managed. It goes without saying that this has substantial societal benefits, since calamities such as the Loveparade, the riots during the Covid-19 pandemic, and the like cause for substantial societal unrest. Moreover, more effective management of crises allows for a more (cost-) efficient deployment of interventions.

### Overall research objectives

The main objective of this research is the development of data-driven predictive risk assessment and decision support methods. To this end, the proposed research entails the following innovations, each of which involves solving several important scientific challenges:

1. Development of **data fusion, state prediction and forecasting methods** using explainable AI and a range of heterogeneous data sources baring information about crowds and crowd dynamics.
2. Development of **risk assessment approaches** that relate the crowd dynamics data with contextual data (weather, sentiments, stress, event characteristics, etc., but also interventions taken) to estimate and predict the risk level and the uncertainty therein.
3. Development and implementation of **novel data collection technology**, including proxies for stress via smart watches for real-time applications and off-line assessment purposes.

These innovations come together in a *Bowtie model* (see Figure 1) linking the innovative data collection methods and readily available data sources to estimate, predict and forecast (crowding) conditions on the one hand, and to assess the (predicted) risks / need to intervene on the other hand.

**Introduction of the CSM project**

Seaside resorts like Scheveningen are usually very pleasant, but when temperature rises and lots of people decide to take to the beach to cool off, a day at the seaside can turn into a nightmare: long traffic jams, far too few parking spaces and huge crowds sauntering down the boulevard. It is under these circumstances that incidents are more likely to occur and the need from crowd management is present.

To facilitate crowd management in Scheveningen, the City of The Hague and regional law enforcement – who are together responsible for crowd management and for safety of visitors, inhabitants, shops, beach club owners in Scheveningen - use a 6-day ahead hourly planning horizon to determine the amount of personnel needed to facilitate crowds coming to Scheveningen. The tactical team planning personnel meets once a week and need a mid-term prediction or forecast 6-days ahead. Until 2022, however, hardly any digital tools were available to support this planning process, and not every team had access to the same information. In the past, the tactical team relied largely on its experiences, taking e.g., school holidays, public holidays, scheduled events, time of year and weather predictions into account when estimating expected crowdedness and related risks 6-days ahead. Because of the structural shortage of law enforcement staff, good crowd predictions are vital. The same personnel can also be deployed at other locations in the region, e.g., at a festival site or in the city centre.

Apart from the 6-day hourly planning horizon used for personnel planning, on each day itself (8-hour, 15-minute horizon), the available personnel need to be at the locations that are most critical or will become critical within the next hour. The operational team determines with areas are most crowded and/or have the highest risk and are together responsible for the allocation of personnel. Most of the information necessary to make these decisions was until recently entirely obtained by people on the ground (walking and cycling through the area). They kept the operational team up to date of what they observed. Again, there were hardly any digital support systems in place. Having the right distribution of personnel over an area or having an early indication that more personnel is needed at one or more spots, potential risky situations can be prevented or managed.

In 2022, Argaleo and TU Delft developed two tools for the city of The Hague and regional law enforcement:

- A common operational picture (DigiTwin) of the crowdedness situation in different areas in Scheveningen. Available open and closed data sources were combined and visualized in such a way that they were easy to comprehend. This *real-time viewer* is a web-based tool, that can be accessed by different people at different locations, providing those responsible for operation and/or planning with the same (visualized) information. If teams wanted to evaluate certain days / events / incidents afterwards, they can search and visualize the relevant *historical data*.
- A short-term prediction (8-hours horizon) and mid-term forecasting (6-day horizon) model to help the operational and planning teams to get predictions of the expected crowdedness and of risks based on that crowdedness. These predictions were included in the Digitwin. In this paper, we focus on the mid-term (multiple day) horizon crowdedness predictions.

In the rest of the paper, we will present the DigiTwin concept as well as the data-driven prediction approaches, focussing on the mid-term forecast. Before doing so, we briefly discuss our overall approach to risk assessment and prediction using a Bowtie framework.

## BOWTIE FRAMEWORK FOR RISK IDENTIFICATION, PREDICTION AND FORECASTING

Assessing risks during events where crowding may occur is not a straightforward task. Next to the volatility of the dynamics of crowds, where high risk situations can develop rapidly, the fact that situations that are similar from a crowd dynamics perspective often have different risk levels. This may for instance be due to different composition of the crowd (e.g., families going to the beach show different behavior than visitors of a dance festival where use of alcohol and drugs may be a factor), the impact of weather conditions (e.g., high temperatures are known to have an impact on the emotional state of people and their cognitive functions), and so on.

The Bowtie model (see Figure 1) aims to structure the factors that influence risk by distinguishing crowd operation factors ("Objective crowdedness") and situational factors ("Aggravating circumstances") that increase (or decrease) the risk given the current (predicted or forecasted) crowding conditions. The left-hand-side and the right-hand-side of the Bowtie are combined to determine the current or predicted risk level. The figure shows several example data on either side of the Bowtie. For the left-hand-side, these include OD data, local density measurements (e.g., from camera footage), weather conditions (affecting demand profiles), parking garage occupancy (providing information about the number of visitors), public transport data, the event calendar and the interventions that have been taken to manage the crowds. For the right-hand-side, the data involves the purpose of the visitors of the event, weather information (as weather influences the sentiments of the crowd), regulations, the number of crowd managers on site, and sentiments and stress of the visitors.

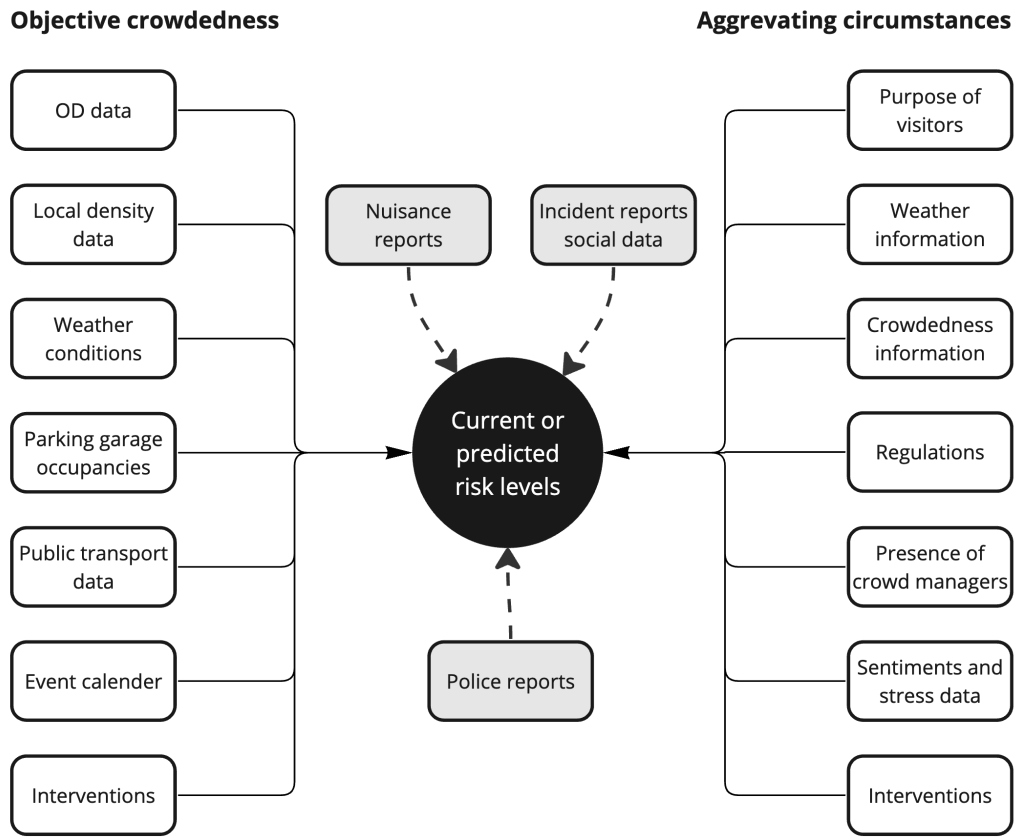

**FIGURE 1 The Bowtie model for risk assessment for data-driven decision support.**

The risk levels identified by the Bowtie are validated by looking at different data sources. Examples in the figure are shown in grey and relate to nuisance reports collected via phone calls, on-line reports via the app, etc., from residents, incidents reported via social media, and police reports).

**State estimation, prediction, and forecasting**

The Bowtie is the conceptual model used in this study. The operationalization of the model is done via application of different interpretable (or explainable) AI methods. For the left-hand-side, we consider state estimation via the fusion of different data sources, on *short-term* prediction and on long-term forecasting. In the context of the application at hand, the *state* refers to those variables that characterise the (objective) situation in our (pedestrian, crowding) network either from the perspective of the decision maker who needs specific information (derived from the variables), or from the perspective of the dynamics of the system (or models thereof) allowing description of the future situation. For the considered domain, this entails variables such as the number of pedestrians in an area, the density, the flowrates, the activity patterns, etc. State-estimation via data fusion entails using a combination of available data sources to optimally estimate the state, which is a classical (and hard) problem in traffic engineering. Previous work by the research team has shown how physics-inspired AI methods are formidable tools to perform this task [8, 9]. This applies equally to short-term prediction of the conditions in the pedestrian network, and to longer-term forecasting: while theory-based models have failed to produce predictions and forecast of sufficient accuracy and reliability, machine learning approaches have been quite successful in fulfilling this task, often at the expense of generalisability of the outcomes. Our ambition is to develop a range of AI and (e)x(plainable)-AI methods [10] that results in accurate estimates, predictions, and forecast of network conditions, including indications of the confidence levels. Distributed *graph neural networks* with *federated learning* seem a promising method [11], since it respects the physical laws of network traffic dynamics, and we can control the complexity of the model and thereby the calibration effort. The distributed training enables keeping data localised if needed, improving data security. In the final step, we will assess the performance of methods.

**Risk identification**

This Bowtie aims to determine the (predicted or forecasted) risk of incidents and calamities, which is used as the main indicator in determining if and which interventions (e.g., getting emergency services in place, re-routing crowds, reduce inflow) need to be prepared or implemented. Incidents entail a range of crowd-related situations that can occur, including dangerous crowding at pinch points, sudden change in weather conditions, but also rioting after a lost soccer match or a demonstration turned sour. State-of-the-practise systems often use trigger values for the different variables or indicators characterising risky situations. However, we argue that risk is a far more complex function of these indicators. Determining this non-linear mapping is far from trivial and is a major challenge.

Like the left-hand-side of the Bowtie, we consider risk classification approaches based on interpretable or explainable AI. Based on a finite number of – possibly multi-dimensional – risk categories, the XAI classifier determines how to relate the estimated or predicted values of the context variables (e.g., crowdedness, weather conditions, sentiments, type of event, measured stress levels) and their uncertainties into a probability (or possibility) that the situation is in a certain risk category. Next to providing accurate classifications, the inferencing method provides interpretable outcomes, showing why a specific situation results in a specific risk. We will start with developing traditional (black box) ANN classifiers. Note that in the remainder of the paper, we focus on the left-hand-side of the Bowtie.

**DATA COLLECTION AND VISUALISATION**

For the CSM, different sources of data are at our disposal. Table 1 provides an overview of the different data sources that are available. Note that for short-term predictions more types of data as well as more accurate data are available than for the multiple day / midterm forecasts.

**TABLE 1 Overview of data sources for the CSM and their role in the Bowtie**

| Name of dataset | Accessibility level | Description | Bowtie side | Horizon |
|---|---|---|---|---|
| Relative crowd information based on mobile app counts (Resono) | Confidential | Historical data visits on different beach areas in Scheveningen and The Hague Area (aggregated to 15 minutes) | Left | Short / midterm |
| Event Calendar Scheveningen | Open access | Dates, times and additional information (ticket sales) about events at Scheveningen Area and public holidays in the Netherlands and surrounding countries | Left | Short / midterm |
| Parking data | Open access | Real-time information on occupancy of public parking garages in Netherlands | Left | Short |
| Public transport data | Open access | Information on public transport schedules, timetables and real-time locations of buses, trains, and trams | Left | Short |
| Shared Mobility data | Open access | Real-time location data of parked shared mobility objects (scooters, bicycles) in the Netherlands | Left | Short |
| Weather information | Open access | Historical, real-time and predicted weather conditions | Right and left | Short / midterm |
| Bicycle counting Netherlands | Open access | Historical and real-time data form bicycle counting systems in the Netherlands | Left | Short |
| Talking traffic data | Shared | Historic trajectory biking data | Left | Short |
| Bridge openings | Open access | Real-time information on bridge openings | Left | Short |
| Floating Car Data | Shared | Travel time data based on floating car data collected from a smart phone app | Left | Short |
| Loop Detector Data | Shared | Speed and flow data from double loop detectors in the network | Left | Short |
| Incident / disturbance reports | Confidential | Information on incidents and disturbances per area | Right | Short |
| Social media data (Twitter) | Harvested | Real-time sentiments determined from social media data | Right | Short |
| Crowdedness estimates / predictions | Confidential | Estimates and short / midterm predictions of crowdedness per area | Right | Short / midterm |

| Crowd management measures | Open access / confidential | Crowd management measures and regulations taken by city or police (signs, rules, road closures, curfew, etc.) | Right | Short / midterm |
|---|---|---|---|---|
| Crowd personnel on streets | Open access / confidential | Number and location of crowd personnel on streets | Right | Short / midterm |
| Purpose of visitors | Confidential | Estimates of visit purposes | Right | Short / midterm |

Due to space limitations, we cannot describe each of the datasources in detail. However, since the Resono data is one of the key datasources used in the ensuing of the paper, we will discuss these in more detail. The Resono crowdedness data is measured by counting the number of smartphones through apps on their mobile panel, consisting of about 2 million users in The Netherlands. Individual smartphone visits cannot be viewed due to aggregation. The University of Groningen and Resono collaborated to validate the existing platform, derived requirements for balancing computational complexity between on-device and cloud computing, and trajectory classification performance. Thorough analysis and scientific standards were followed for robustness and performance assessment. Additionally, a rigorous empirical evaluation used historical and 'ground truth' datasets representing realistic scenarios for validation of strategies and models; see https://reso.no/certificeringen-en-validaties/ for more information.

For CSM, the Resono data is collected for 16 areas at the Scheveningen beach site. For each of the areas, the (relative) number of visitors can be determined per 15 minutes. Moreover, the data shows movements from area to area, to areas in and outside The Hague, as such providing origin-destination information about the visitors. An example is shown in Figure 2. Next to the Resono data, for the results presented in this paper we consider weather data (historic, current and forecasted) including the temperature, the actual precipitation, the probability of precipitation, cloudiness, and windspeed.

**Digital Twin environment**

For the CSM, the company Argaleo has built a dedicated 3D DigiTwin in which all kinds of standard and non-standard data services are linked that can be visualized, filtered, and analyzed via a simple menu. Depending on focus and zoom level, the DigiTwin shows the underlying maps and information in 3D. By integrally displaying various linked data sources, the user gets a coherent overview of a specific area. Analyses can be performed that provide an overview of the necessary indicators for specific areas or objects. For example, crowdedness in an area, number of open parking spots in a parking garage, location of available shared mobility, and locations of buses, trams, and trains; all (near) real-time. An impression of the DigiTwin is shown in Figure 2.

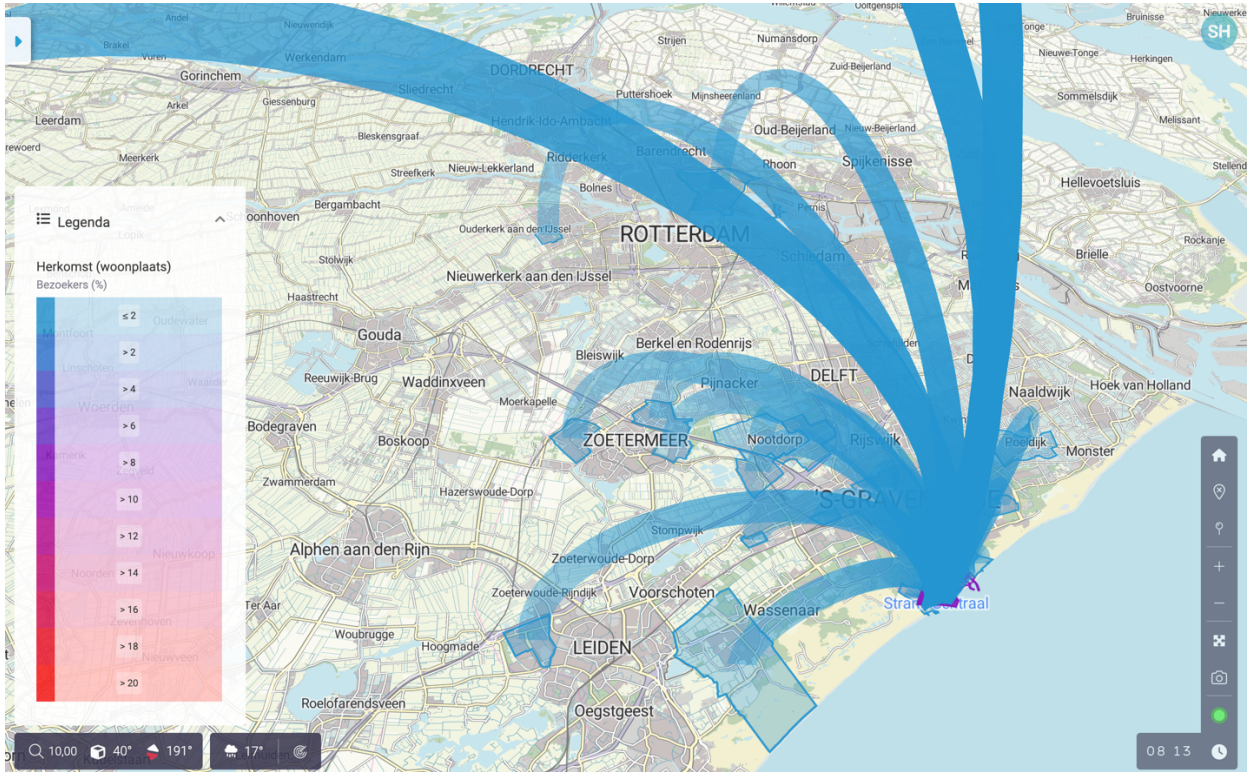

**FIGURE 2 Example of the DigTwin showing Origin Destination information of the visitors of one of the 16 identified areas in Scheveningen.**

**Descriptive statistics of relevant data**

For illustration purposes, let us present some of the data present in the CSM system. As there are many different datasources in the DigiTwin, we had to select the sources presented.

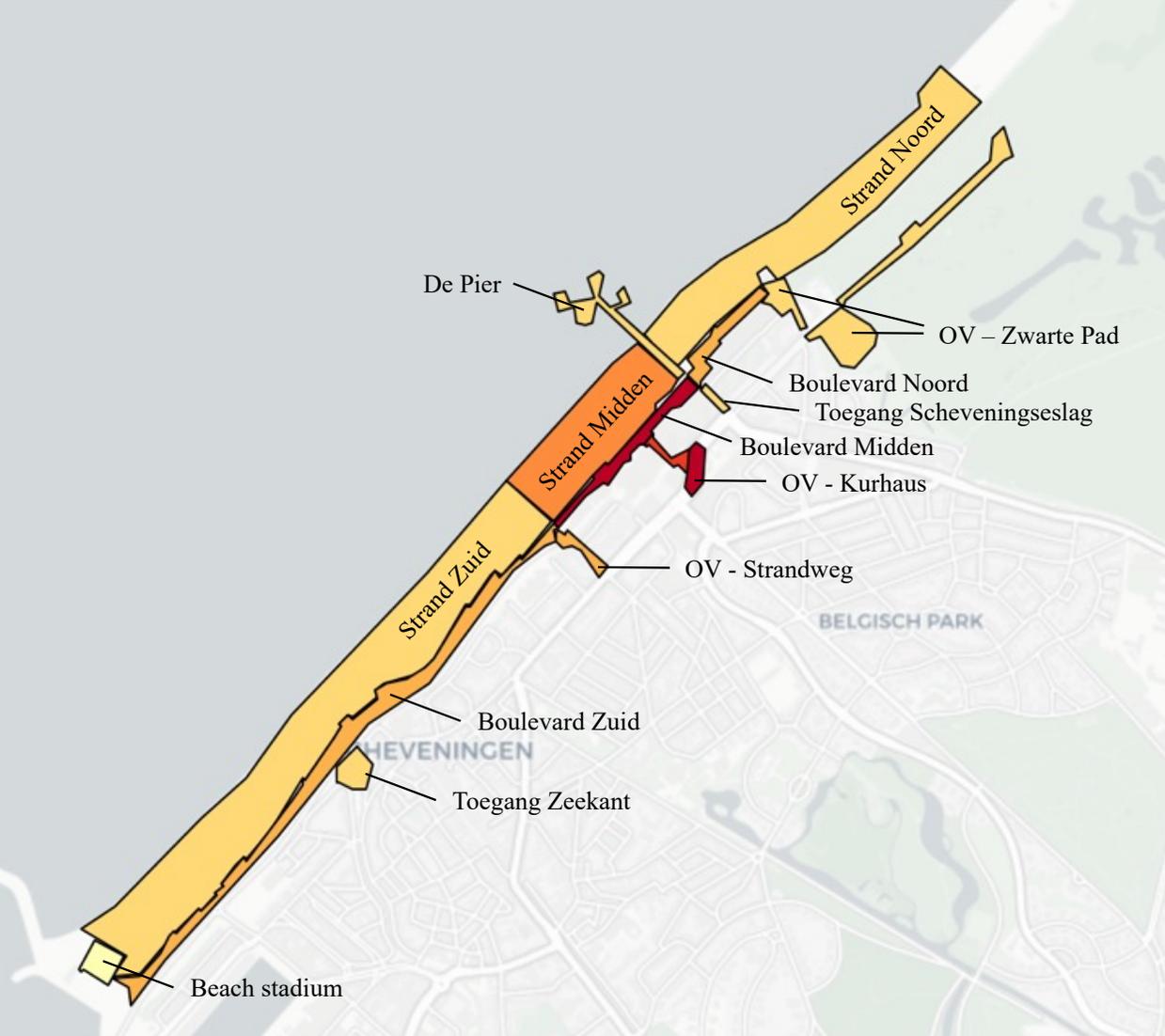

**FIGURE 3 Overview of areas and their labels**

Figure 4a shows the number of visits in the 16 zones in Scheveningen, depicted in Figure 3, for the period of April 2021 to May 2022. We emphasize that these different locations have different functions. For instance, the Boulevard Noord / Midden / Zuid and De Pier areas are walking zones with different shops and restaurants; strand Centraal / Noord / Zuid are the main beach areas. The Beach Stadium is a location where events are organized. The different "toegang" locations are access locations to specific areas; OV Kurhaus / Strandweg / Zwarte Pad are locations including public transport stops. While the functions of these areas are quite different, the average distributions of the visitors over the day are quite similar (Figure 4b).

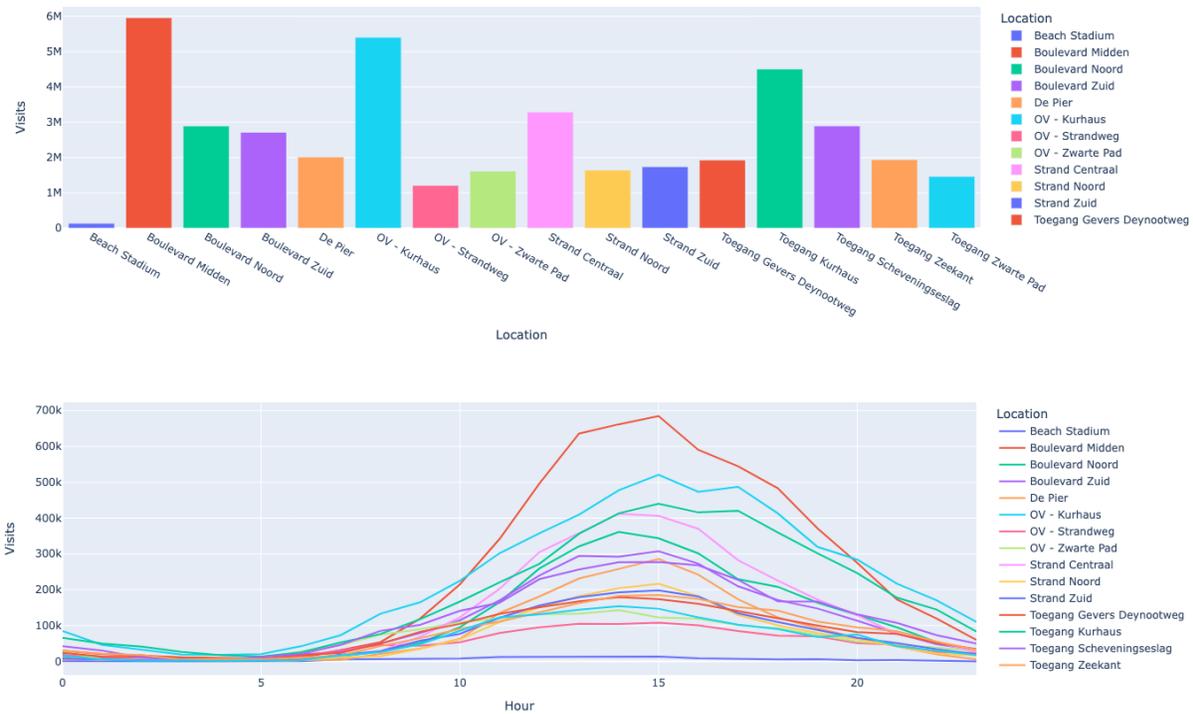

**FIGURE 4a) Total number of visitors for the period April 2021 to May 2022; b) Distribution of the visits over the day per zone.**

Figure 5a shows the number of visitors for one of the 16 zones, De Pier, in comparison to the temperature (5b) and the wind (5c). These data form the starting point for the preliminary analysis performed before building prediction models using ML.

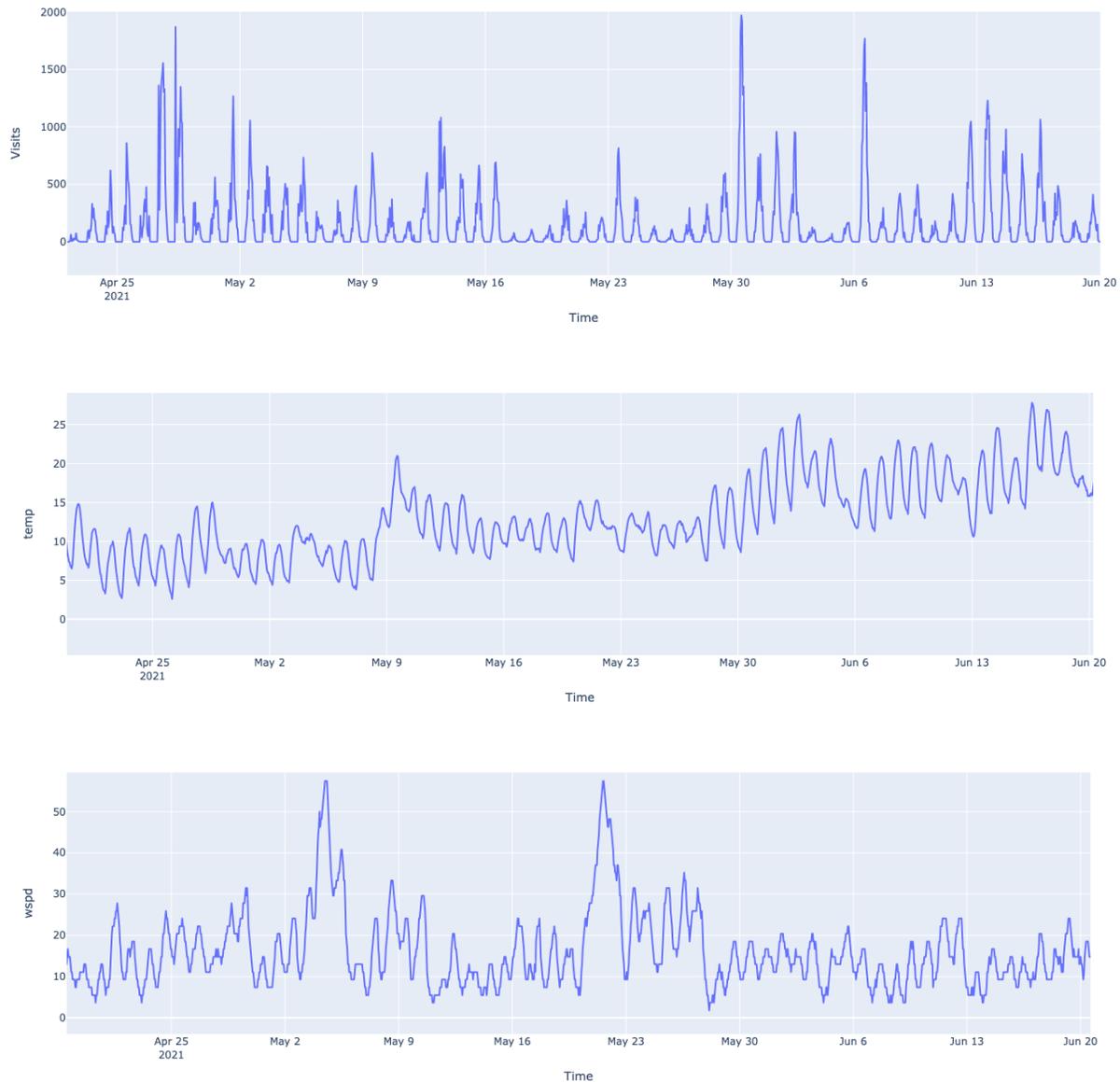

**FIGURE 5a) Hourly visitor pattern for De Pier; b) Hourly temperature pattern for Scheveningen area; c) Wind speed pattern for Scheveningen area in April / June 2021.**

To gain some first insight into the relations between the datasources at our disposal, we performed some simple multivariate regression analysis to see which factors influence the deviations from the mean number of visitors. It turned out that, next to the historic trends in the visitor number per area, mostly weather-related indicators had a significant impact on the number of visitors. For example, we looked at the difference between the average number of visitors shown in the picture above and the actual number of visitors; see Table 2.

**TABLE 2 Multi-variate regression results for total number of visitors (aggregated over all areas) and some weather-related factors. All factors are significant at 95% confidence.**

| Attribute | Mean b | 5% value | 95% value |
|-----------|--------|----------|-----------|
| Temperature | 225 | 212 | 238 |
| Prob precipitation (0-100%) | -10 | -13 | -7 |
| Clouding | -29 | -32 | -27 |
| Windspeed | -290 | -326 | -254 |
| Weekend (=1) or not? | 3246 | 3029 | 3464 |

Further investigations into the correlation between temperature, wind speed and number of visitors does reveal a clear non-linear relationship. Therefore, in the remainder we mainly look at non-linear ML methods.

## FORECASTING CROWDEDNESS USING ARTIFICAL INTELLIGENCE

Here, we will focus on forecasting methods for the crowdedness in the different areas in Scheveningen $M$ days ahead, with the aim to support crowd management planning. We do so by using forecasted weather including windspeed for the next $M$ days and the visitors' data for the previous $N$ days. We expect that the proposed methods will be applicable to short-term prediction problems as well but will not discuss this in detail here. Below, we will discuss the different steps of the approach, including feature selection, on-step ahead prediction, multi-step ahead prediction, and evaluation.

### Feature selection

Feature selection is a crucial step in our data preprocessing pipeline, and it heavily influences the performance of our machine learning model. It is also important in helping us understand the structure and relevance of our data. In our dataset, we extracted temporal features from the data such as month, weekday and hour and these were included in the full feature set. We used the filter method as a pre-processing step. We calculated the correlation between each feature and the target variable to select the most relevant ones, like the multi-variate regression shown in the previous section. Features that demonstrated low variance or no correlation with the target variable were removed, as these likely contributed more noise than meaningful information. This not only helped us reduce the dimensionality of our dataset but also accelerated our model training process and mitigated the risk of overfitting.

The insights derived from these correlation measurements also informed our decision on whether to use linear or non-linear machine learning methods. A strong linear correlation between the selected features and the target variable suggested that linear methods like linear regression might perform well. However, if any non-linear relationships were inferred, non-linear methods such as decision trees or random forests might be more appropriate.

The results of the feature selection process, stemming from both the initial selection and from analysis of the outcomes of the one-step ahead prediction using the XGBoost feature importance discussed in the next section, is shown in Figure 6.

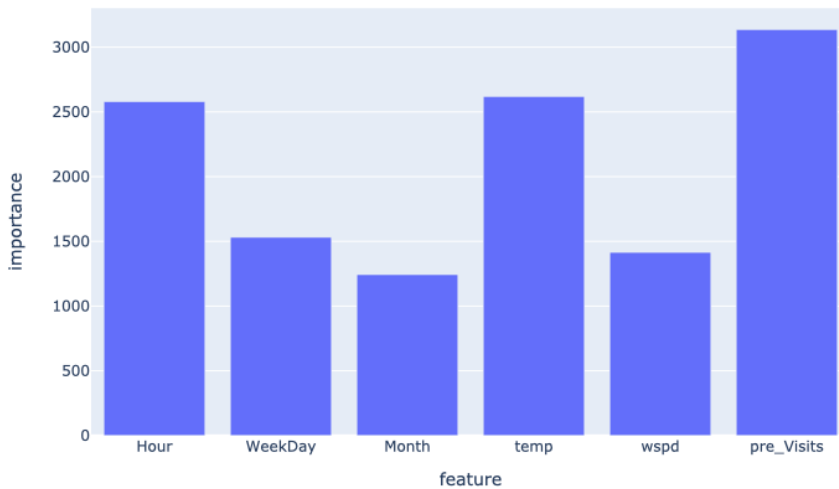

**FIGURE 6 Results of the feature selection process, showing the main features considered (based on their F-score), including hour of the day, day of the week, month of the year, temperature forecast, windspeed forecast, and number of previous visits.**

**One-step ahead prediction**
In this study, we employed a linear and non-linear machine learning method to develop predictive models focusing on predictions (or rather, forecasts) one or multiple days ahead. The specific methods used are Multiple Linear Regression and XGBoost. Each method was selected based on its unique strengths and was evaluated based on its performance for one-step ahead prediction on a validation set.

*Multivariate Linear Regression as Benchmark Method*
As already indicated in the previous section, Multivariate Linear Regression (MLR) is a powerful statistical method used to understand the relationship between two or more features and the target variable. This method served as a benchmark for comparing the performance of more complex models. The main parameters in multiple linear regression are the coefficients of the predictors, which are learned during the training process, referred to as the coefficient of determination $R^2$. However, we also considered the significance level (alpha) for the F-test to determine whether the coefficients are statistically significant.

*XGBoost*
XGBoost, short for Extreme Gradient Boosting, is an ensemble learning method where multiple weak learners, decision trees, are combined to create a more accurate and robust model [12]. XGBoost is a sequential learning method where each new tree is built to correct the errors made by the existing ensemble of trees. This is different from methods like Random Forest, where each decision tree is built independently. It is effective in a wide range of tasks, especially structured/tabular data problems. There are several parameters that need to be tuned for XGBoost. The most important ones are the maximum depth of the tree which helps control overfitting and the number of gradient-boosted trees.

One important property of tree-based methods is their capability to derive feature importance. The metric used for feature importance is the F-score which represents the number of times a feature is used to split data across all trees. A higher F-score implies greater importance and indicates the feature's usefulness in decision-making within the model. This score can be used for feature selection, improving model interpretability, and understanding the relationships between features and the target variable. The results of this step is shown in Figure 6.

**Multi-step ahead prediction**

Our study addresses long-term multi-step ahead prediction, an essential aspect of forecasting in time series analysis. Given the increased potential for error propagation in long-term forecasts, we concentrated on the two strategies: the direct and recursive methods [13]. The direct strategy involves developing distinct models for each forecast step. While this approach allows each model to be fine-tuned for its specific future time step, it can become computationally heavy as the prediction horizon extends, making it less appropriate for long-term predictions.

The recursive strategy uses a single model to make predictions one step ahead, which are then fed back into the model to generate further predictions. While this method is more computationally efficient, it faces the challenge of error accumulation over time, which can diminish the accuracy of long-term forecasts. In this study, we carefully assessed the trade-offs between these strategies, considering factors such as computational resources and forecast accuracy, and chose the recursive strategy for the multi-step ahead prediction.

**Evaluation**

To assess the performance of our models, we utilized several evaluation metrics, specifically Mean Absolute Error (MAE), Root Mean Square Error (RMSE), and Mean Absolute Percentage Error (MAPE). Each metric provides a different perspective on the model's error, and their combined use gives a more comprehensive evaluation of model performance. For instance, the MAPE, and to a lesser extent MAE, is less effective when dealing with target variables that take very small values since a small absolute error can turn into a large percentage error if the actual value is very small, leading to an overestimation of the error size. Similarly, MAE may not reflect the true error magnitude when the target variable values are low. For evaluating the multi-step ahead prediction, we computed the absolute error for each one-step-ahead prediction and then averaged the errors.

## FORECASTING MODEL DESIGN AND RESULTS ANALYSIS

In this section, we present the main results of the forecasting algorithm developed. We will discuss the outcomes of the XGBoost in detail, as our cross-comparison showed that it performed better than the MLR in terms of accuracy, while at the same time resulting in interpretable models via the regression trees as shown in Figure 7. All MLR models have a $R^2$ greater than 0.8, ranging from 0.8 at Beach Stadium to 0.93 at OV Kurhaus. Compared to linear regression, XGBoost improves the MAE by an average of 24%, with a maximum improvement of 44% at Boulevard Midden. Locations with the most visits, like OV Kurhaus and Toegang Kurhaus, saw improvements of 24% and 30% respectively.

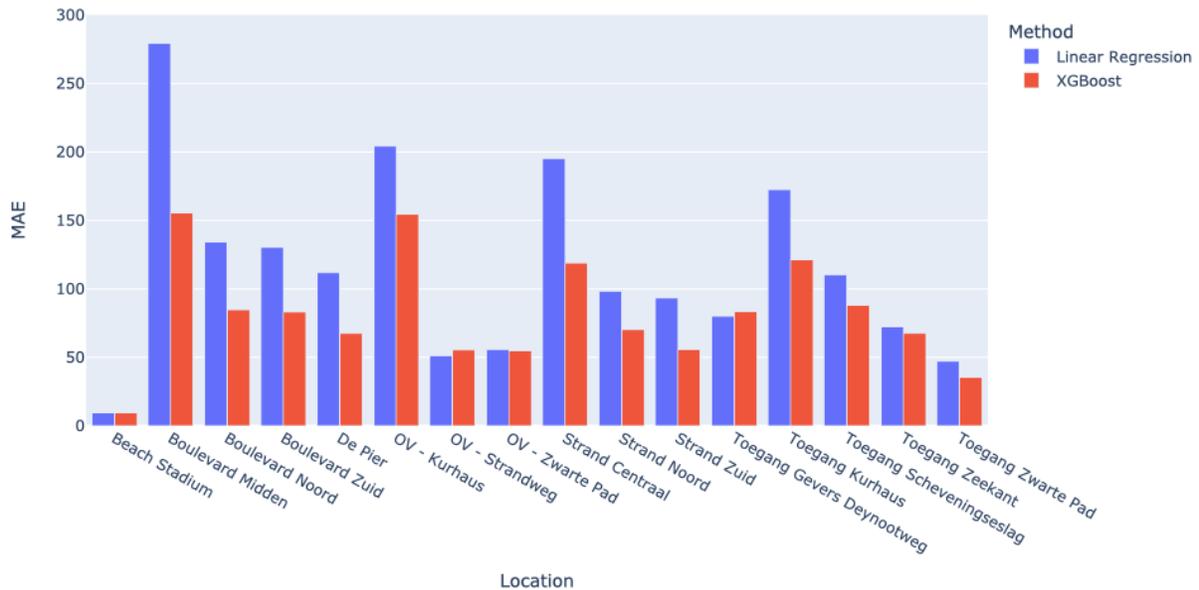

**FIGURE 7 Comparison between multi-variate regression and XGBoost per Scheveningen area.**

In the remainder of the section, we discuss the results of the one-step-ahead prediction, starting with the design of the model. We then show the results of the one-step-ahead prediction in terms of MAE analysis for the different areas considered. We subsequently briefly discuss the multi-step ahead predictions for some of the areas.

**Design of the gradient boosting framework**

The most important design parameters of XGBoost are the depth of the tree and the number of estimators. The number of estimators refers to the number of individual weak learners (base models) that are sequentially combined during the training process. Each estimator is a decision tree, which added to the ensemble in an iterative manner to improve the model's predictive performance. Increasing the depth and the number of estimators can make the model more complex and potentially improve its predictive performance. However, it also increases the risk of overfitting, where the model becomes too specific to the training data and performs poorly on new, unseen data. To prevent overfitting, it is essential to tune the number of estimators and use techniques like early stopping during the training process.

Figure 8 shows the impact of tree depth and the number of estimators (trees) on the model performance. It clearly shows that beyond a tree depth of 7, no major improvements are observed. For the number of estimators, we see that for 11 estimators or more, no clear improvements are observed. In the remainder, we, therefore, choose a maximum depth of 10 with 15 estimators.

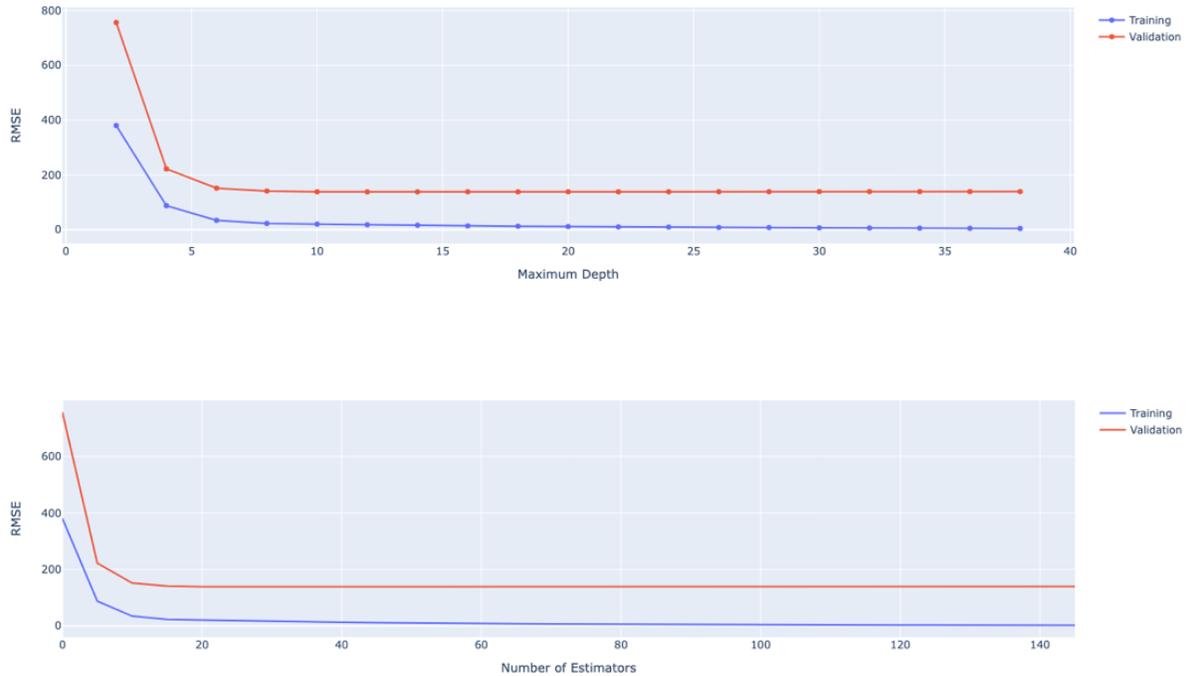

**Figure 8a) Impact of regression tree depth on the RMSE; b) Impact of the number of estimators on the RMSE.**

### One-step ahead prediction results

To get insight into the quality of the one-step ahead predictions, Figure 9a compares the ground truth (Resono counts aggregated to one day) to the one-step (i.e., one day) ahead predictions. From the limited scatter in the figure, we conclude that in general the forecasts provided by the model are of sufficient quality. We are aware that what is "sufficient" depends on the application of the predictions, which in this case is providing information to the key stakeholders (policy, crowd managers, event organisers) for planning purposes. The authors were in regular contact with the technical team and heard first hand that the mid-term forecasts were accurate enough to support their decision-making tasks. Figure 9b shows the error for the different areas in Scheveningen.

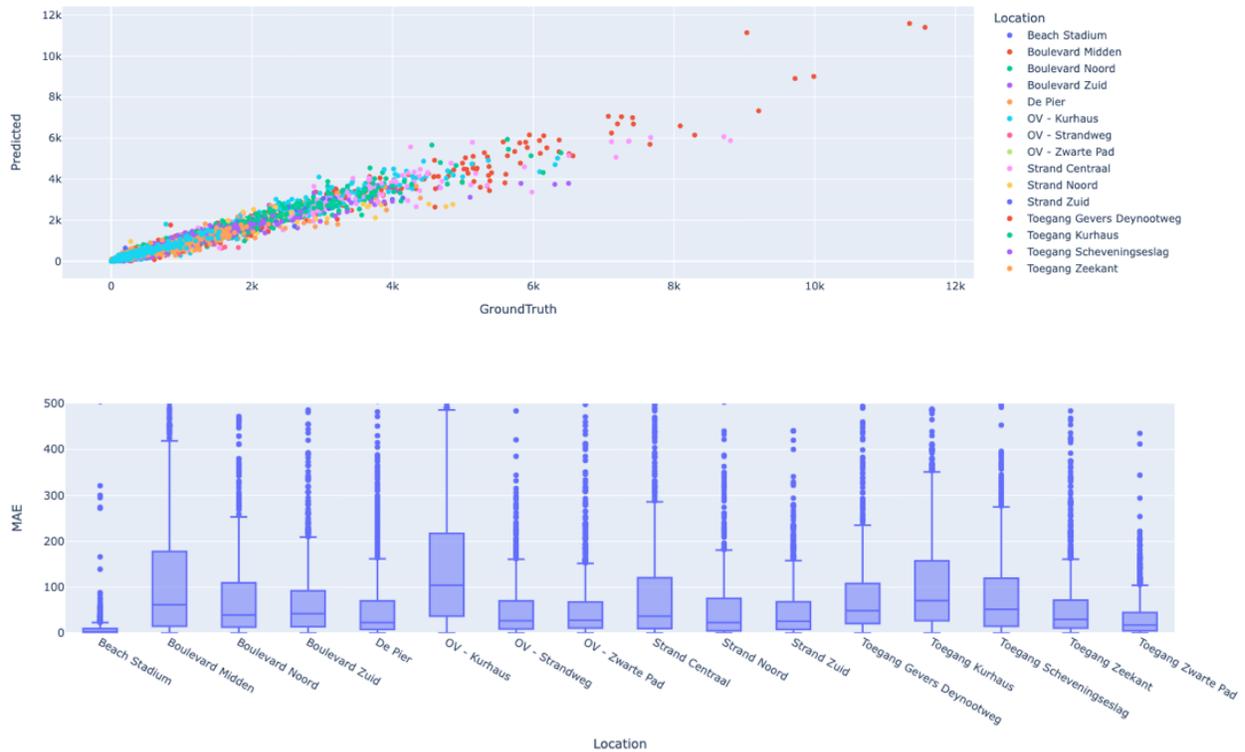

**Figure 9a) Comparison between ground truth and one-step ahead prediction; b) boxplots of error (MAE) for the different Scheveningen areas.**

**Multistep-ahead prediction results**

As a final analysis, we briefly look at the multi-step (or rather, multi-day) ahead prediction. Figure 10 shows the average error of the 10 days multi-step ahead prediction. In comparison to Figure 9b, we can observe an increase in the error that is due to the error accumulation over time due to recursive strategy.

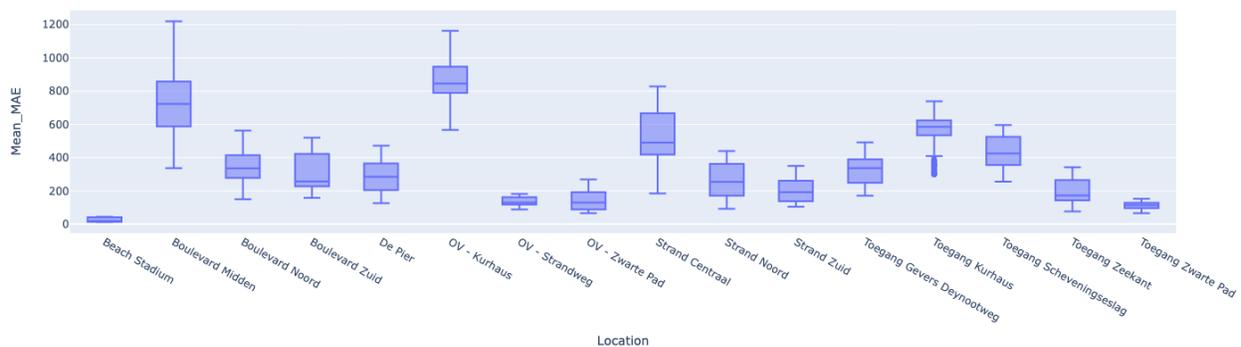

**Figure 10 Comparison between ground truth and one-step ahead prediction; b) boxplots of error (MAE) for the different Scheveningen areas.**

To provide further insights, below we highlight the outcomes of four specific locations (Figure 11): De Pier, De Kurhaus (which has the highest absolute prediction error), the Beach Stadium (which has the lowest number of visitors), and the Boulevard (which has the highest number of visitors). For each of the locations, we have predicted 10 days ahead, starting from Saturday, 9th of April 2022. In general, the figures show that the trends are very well captured, and the daily trends are predicted very well. Errors appear to be larger for the predictions that are multiple days ahead.

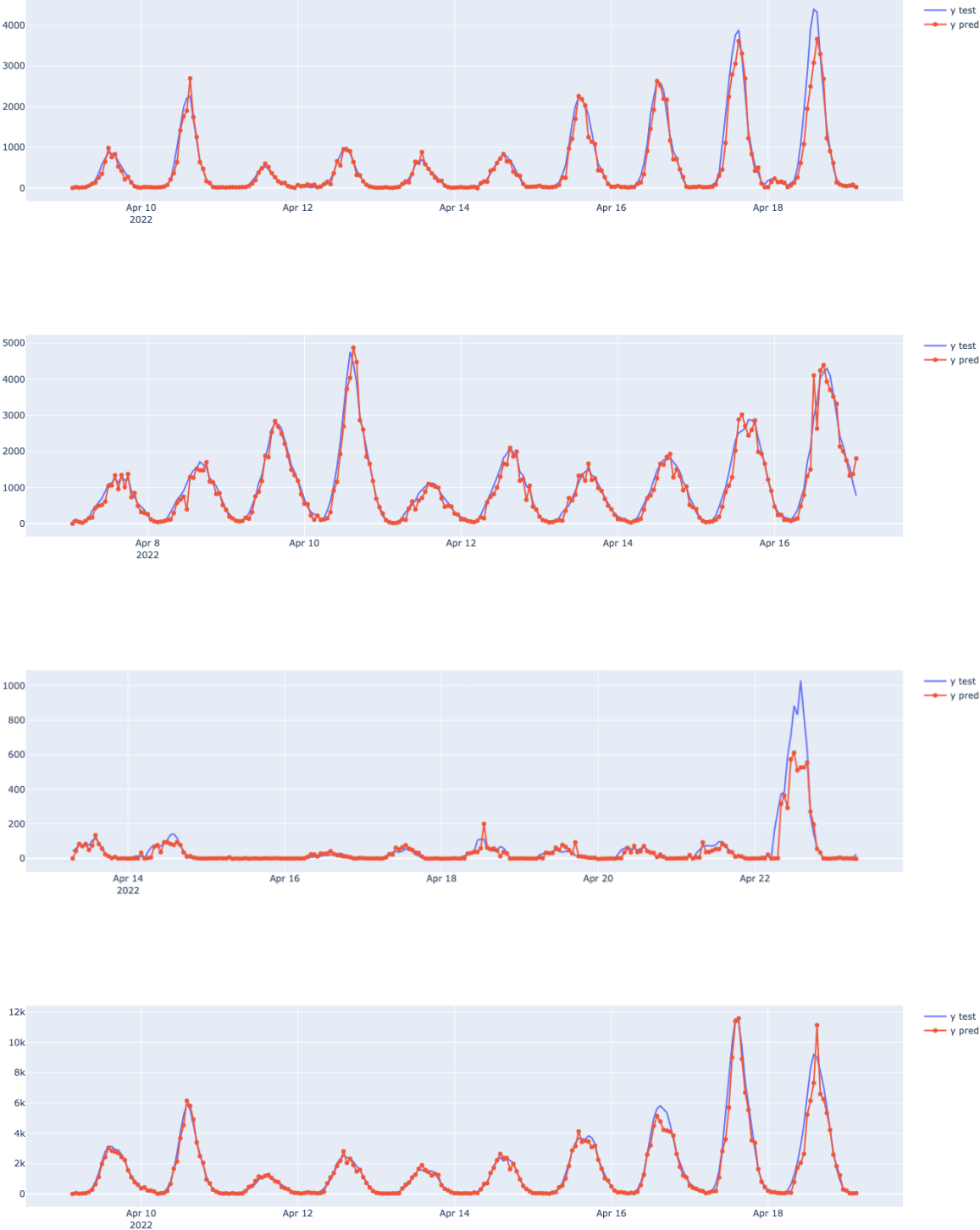

**FIGURE 11 Ten-step ahead prediction examples for a) De Pier; b) De Kurhaus (zone with the highest error); c) Beach Stadium (zone with lowest visitors); d) Boulevard (combination of zones with highest number of visitors).**

The relative largest error is made at the Beach Stadium area. This is expected, as this area is generally only crowded if events are organized there and are less influenced by weather conditions since most event require a pre-sold ticket. To remedy this, in the future additional data sources (i.e., the event calendar) will be added to improve the predictions.

## CONCLUSIONS AND OUTLOOK

In this final section, we present our main conclusions. Moreover, we discuss the limitations of our study and provide directions for future research.

### Findings and conclusions

This paper presents novel technology and methods for supporting crowd management at both the planning and the operation phase. The approach comprises of novel data collection methods, data integration and visualisation using a 3D Digital twin, as well as AI tools for risk identification. For the latter, the paper presented the Bowtie model, proposed to assess current and predicted risk levels. The Bowtie combines objective (traffic flow operation) estimations and predictions (e.g., crowdedness levels, origin-destination flows) with aggravating circumstances (weather conditions, sentiments, purpose of visitors, etc.) to assess the (expected) risk of incidents.

The presented framework is applied to the Crowd Safety Manager project in Scheveningen, for which the DigiTwin is developed based on the large number of (real-time) data sources available. One of the innovative data sources available is Resono, which provides information on the number of visitors and their movements, based on a mobile phone panel of over 2 million users in the Netherlands. In the project, we furthermore focussed on the left-hand-side of the Bowtie (state estimation, prediction, and forecasting), and in particular on multiple-day ahead forecast to be using these Resono data for personal planning purposes. After a multi-variate regression analyses, advanced machine learning methods were developed and cross-compared. As the XGBoost framework resulted in the most accurate forecasts, we continued with detailing its design (feature selection, regression tree depth, number estimators) of and analyses the one-day and multi-day ahead predictions.

Overall, it turned out that the predictions were sufficiently accurate. For some specific locations, however, additional input data is likely needed to further improve the prediction quality: while the considered features (previous visits, temperature forecast, wind forecast) were suitable for most locations, for some additional information (e.g., event calendar) is likely required to improve prediction accuracy. That said, we can conclude that the proposed technology and methods are very promising and provide important support for planning events where crowding may potentially become an issue.

### Discussion and outlook

The presented work has several limitations. First, short-term predictions (e.g., 15 minute or 1 hour ahead) have not been considered. For operational decision support, these are very important. We expect that the presented methods are suitable for these short-term predictions. We argue that for short-term predictions more timely and accurate data are available than for the multiple day ahead forecasts. On the one hand, this might complicate our short-term prediction model. On the other hand, the use of 'more recent' data (e.g., one day instead of 6 days weather prediction) as well as the availability of a larger variety of data sources (PT data, shared mobility data, road network intensities and speeds, bicycle counts, and parking garage occupancies) will likely improve predictive capabilities. That said, the accuracy of short-term predictions on a busy beach day will be more critical. Due to aggravating circumstances, e.g., changes in sentiment of part of the beach crowd, a mixture of beach visitors with differing motives, or rapidly changing weather conditions, makes risks more volatile and increases the need to be more proactive to prevent or manage risks.

Second, the paper focussed on the left-hand-side of the Bowtie. While crowding is one of the key factors in risk assessment, other factors will most likely also influence the likelihood of an incident.

Future work will focus on empirically underpinning the impact of the factors present at the right-side of the Bowtie and combining them with the factors on the left side. The use of interpretable AI technology (e.g., regression trees) will allow gaining understanding in the (non-linear) combinations of factors on each side of the Bowtie in explaining, assessing, and predicting risks.

As a final point, we need to be aware that decision-making for crowding incidents is characterized by high levels of uncertainty, dynamics, and time pressure. Decisions to manage crowding combine the time-pressure typical for operational crisis management with the complexity and novelty of high-stake decisions. Earlier research has shown that in such situations, decision performance and situational awareness are sensitive to the role and cognitive load of the decision-maker [14] and that initial information and sensemaking create rigid mental models that are not adaptive to new information [15]. In the future, we aim to formalize the role of information for decision support in expert decision making and investigate information-decision-feedback.

## ACKNOWLEDGMENTS

The crowd safety manager is an initiative of the National Police and the municipality of The Hague. In cooperation with partners TU Delft, Q-TC Nederland B.V., TNO, WE LABS, Future City Foundation, ELBA\REC and Argaleo. The Netherlands Enterprise Agency (RVO) provided a grant for the development of the crowd safety manager. We would like to acknowledge the utilization of OpenAI's chatGPT-4 in refining the text in the methodology section.

## AUTHOR CONTRIBUTIONS

The authors confirm contribution to the paper as follows: study conception and design: Krishnakumari, Hoogendoorn, Hoogendoorn-Lanser; data collection: Steenbakkers, Hoogendoorn-Lanser; analysis and interpretation of results: Krishnakumari, Hoogendoorn; draft manuscript preparation: Krishnakumari, Hoogendoorn, Hoogendoorn-Lanser. All authors reviewed the results and approved the final version of the manuscript.